\documentclass{article}

\usepackage{threeparttable}
\usepackage{booktabs,caption}
\usepackage{arxiv}

\usepackage[utf8]{inputenc} 
\usepackage[T1]{fontenc}    
\usepackage{hyperref}       
\usepackage{url}            
\usepackage{booktabs}       
\usepackage{amsfonts}       
\usepackage{nicefrac}       
\usepackage{microtype}      
\usepackage{lipsum}
\usepackage{graphicx}
\graphicspath{ {./images/} }

\title{Topology-Driven Generative Completion of Lacunae in Molecular Data }

\author{
 Dmitry Yu. Zubarev \\
  IBM Research - Almaden\\
  650 Harry Rd.\\
  San Jose, CA 95120\\
  \texttt{dmitry.zubarev@ibm.com} \\
   \And
 Petar Ristoski \thanks{Current email:  pristoski@ebay.com}\\
  IBM Research - Almaden\\
  650 Harry Rd.\\
  San Jose, CA 95120\\
  \texttt{petar.ristoski@ibm.com} \\

}

\begin{document}
\maketitle
\begin{abstract}
We introduce an approach to the targeted completion of lacunae in molecular data sets which is driven by topological data analysis, such as Mapper algorithm. Lacunae are filled in using scaffold-constrained generative models trained with different scoring functions. The approach enables addition of links and vertices to the skeletonized representations of the data, such as Mapper graph, and falls in the broad category of network completion methods. We illustrate application of the topology-driven data completion strategy by creating a lacuna in the data set of onium cations extracted from USPTO patents, and repairing it. 
\end{abstract}


\section{Introduction}

Materials discovery is frequently driven by historical data sets that lack characteristics of the data sets specifically constructed to meet the needs of particular discovery efforts. They carry imprints of the ever-changing historical context of the research and development. Shifting priorities of the external funding, pressure for momentous technological breakthroughs, community perception of high-profile topics, and evolution of experimental capabilities render historical data a patchwork of findings with poorly understood internal structure. Statistical learning methods are typically concerned with statistical characteristics of the data. In the materials discovery, there is an additional pressure to understand the shape of the data in terms of what is known and what is missing and inform laborious and expensive data acquisition associated with material preparation, processing, and characterization. 

In this contribution, we are investigating the interplay between the shape of the historical data expressed as the structure of lacunae, such as gaps, loops, and voids, and the hypothesis generation that informs subsequent data acquisition. We describe an approach that explicitly identifies lacunae via topological data analysis (TDA) and fills them in using constrained generative modeling. TDA is concerned with capturing the shape of the data - the characteristics that are preserved under continuous deformations. The simplest widely accepted form of TDA is clustering. Development of the efficient computational algorithms in combination with ever-growing computational power catalyzed the adoption of more engaging and informative forms of TDA, including persistent homology (PH) \cite{PH:2013, PH:2017}, and Mapper TDA \cite{TDA:1, TDA:2, Mapper:1, Mapper:2, Mapper:3, Mapper:4, Mapper:5, Mapper:6}. Both PH and Mapper reveal lacunae in the data: persistent homology does so in the exhaustive manner and focuses on capturing the scales where lacunae exists; Mapper constructs sckeletonized representation of the data set that corresponds to a particular scale and reveals lacunae that might exist on that scale. 

Generative modeling and a broader field of inverse design gained traction in materials discovery with the hope to alleviate challenges of brute-force approaches such as combinatorial design and high-throughput screening. Deep generative models were first introduced in the computer vision field \cite{pan2019recent}, and can be divided in three categories: Generative Adversarial Networks (GANs) \cite{goodfellow2014generative}, Variational Autoencoders (VAE) \cite{kingma2013auto} and AutoRegressive Networks \cite{akaike1969fitting}. Such approaches were quickly adapted in other research fields, such as molecular design, which significantly improved the efficacy of generating novel molecules \cite{sanchez2018inverse, elton2019deep, schwalbe2020generative}. 

The historical data set explored in this study comprises sulfonium and iodonium cations extracted from USPTO patents (see ``Data preparation'' for the details of the data set construction). Sulfoniums and iodoniums, as members of a broader family of oniums \cite{OniumGen:1997}, appear in multiple chemical applications due to their reactivity, including photo-initiated transformations \cite{OniumRev:2020, PAGReview:2020}. The capability of sulfoniums and iodoniums to generate strong acids in response to the exposure to UV radiation has been a driving force of the chemical amplification technology of lithography that shaped up semiconductors production \cite{OniumPhoto:1992, OniumPhoto:1999, SulfoniumPhoto:2009}. There is a continued interest to identify new members of sulfonium and iodonium families with superior performance in lithogrpaphy, favorable environmental properties, and potential for intellectual property generation.

\section{Data, Methods, and Computational Details}
\label{sec:methods}

\begin{figure} 
    \centering
    \includegraphics[width=1.0\textwidth]{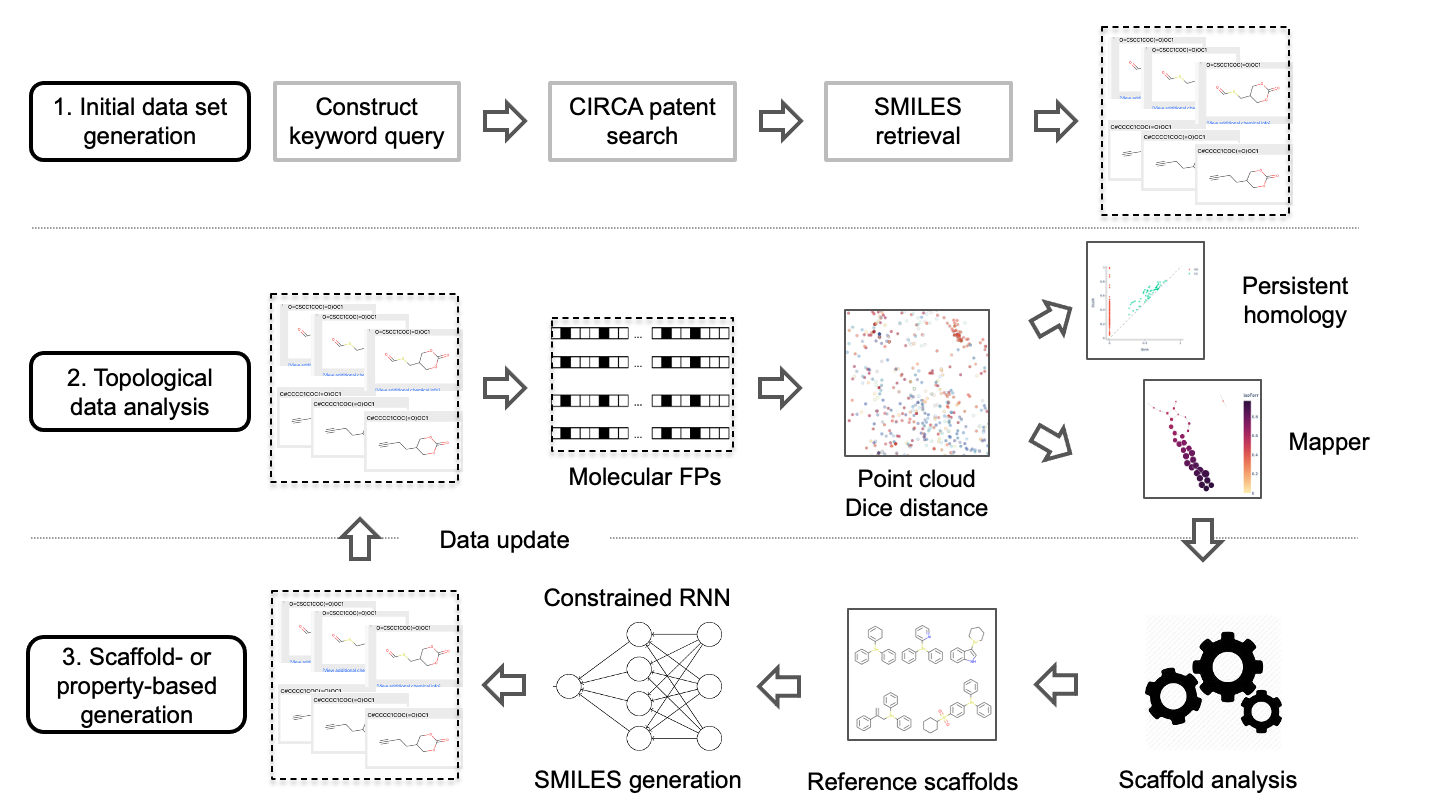}
    \caption{\label{fig:workflow} Schematic representation of the computational workflow of topology-driven generative completion of molecular data.}
\end{figure}

The overarching motivation of our choice of methods and codes in this study was to maximize re-use and emphasize interaction between mature computational technologies that have already propagated into chemical science domain. In this manner, we hope to achieve a faster adoption of the proposed composite approach to the data completion. We aim to a) transform TDA from a descriptive tool into a source of actionable insights, and b) achieve a higher degree of control over the outcome of generative modeling. The computational workflow of our study is outlined in Figure \ref{fig:workflow}. Stage 1 ``Initial data set generation'' is described in subsection \ref{subsec:data}; details of stage 2 ``Topological data analysis'' are provided in subsection \ref{subsec:tda}; finally, subsection \ref{subsec:generative} describes stage 3 ``Scaffold- or property-based generation''.

\subsection{Data preparation}
\label{subsec:data}
Sulfonium and iodoniums cations were harvested from Chemical Information Resources for Cognitive Analytics (CIRCA) \cite{CIRCA} that hosts a comprehensive corpus of the annotated USPTO patents. The search was performed for the keywords ``sulfonium'' or ``iodonium'' that are accompanied by chemical annotation in the patent texts. The search returned approximately 5.5K items that appeared in the patents filed between 1976 and 2019. We expect to capture a technologically relevant set of cations due to the fact that their names are explicitly mentioned in the patent texts. Some ``contamination'' of the data set is anticipated due to cations that do not constitute intellectual property on their own and found their way into the patents as components of relevant chemical processes. We also recognize that the keyword-based search did not return cations described as drawings or included in accompanying materials.
SMILES representations of the cations were processed to remove symbols encoding chirality and cis/trans-stereochemistry. The distances between cations were computed as Dice distances on Morgan topological fingerprints in the bit-vector form. All chemoinformatic manipulations were performed using \textit{RDKit} library\cite{rdkit}. We analyzed the data set for the presence of outliers by computing anomaly scores of the cations per Isolation Forest algorithm as implemented in \textit{scikit-learn} library\cite{scikit-learn}. In Isolation Forest anomaly detection, the lower values of the decision function are indicative of higher abnormality of the data points. The lowest encountered anomaly score in the compiled data set is 0.04 and the highest anomaly score is 0.20; there are no clear outliers with negative scores. 

\subsection{Topological Data Analysis: Persistent Homology and Mapper}
\label{subsec:tda}
TDA is conceptually rich and can be computationally involved \cite{TDA:1}. Multiple tutorials offering technical descriptions of both PH and Mapper at various depth can be found on-line. We prefer to avoid detailed technical discussion of either method in order to maintain the focus of the paper; in this section we describe the practical aspects of PH and Mapper analyses that are required to understand the reported results. 

As a form of TDA, PH \cite{PH:2017} computes topological features of the data set at different spatial resolutions. The theoretical foundation of PH is algebraic topology; recent progress in the development of efficient software packages tremendously simplified application of PH analysis in chemical domain \cite{PHMaterials:2020}. A topological feature can be a connected component associated with disappearing gaps, 1D hole/loop, 2D cavity, or a higher-dimensional “void” which exists in the data. We captured these features via computation of Vietoris-Rips persistence \cite{PH:2017}. Results of the computations were expressed as a persistence diagram, where each identified topological feature was assigned two values: the \textit{birth} value, characterizing the length scale at which the feature appeared,and the \textit{death} value, characterizing the length scale at which it disappeared. The difference between \textit{death} and \textit{birth} values determined the \textit{lifespan} of the feature. The computational demands of PH calculations noticeably increase with the dimensionality of the features. We computed Vietoris-Rips persistence for the features of degree 0 and 1, designated throughout the text as \textit{$H_0$} and \textit{$H_1$}, respectively. Computations of Vietoris-Rips persistence were performed using \textit{giotto-TDA} library \cite{gTDA:2021};  

Mapper TDA combines concepts of Reeb graph \cite{Mapper:0} and pullback of a covering map. It offers a direct visualization of topology on complex high-dimensional data without dimensionality reduction. Mapper tracks evolution of a real-valued function, a \textit{filter function} or a \textit{lens}, associated with the data points. We selected Isolation Forest anomaly score as the filter function. The range of the filter function values was split into 30 overlapping intervals (50\% overlap). The cations in the molecular point cloud that fell into the same intervals according to their anomaly score were assigned to the same \textit{level sets}. Each level set was clustered to reveal disjoint subsets in it. We used spectral clustering algorithm (\cite{scikit-learn} implementation) with 2 target clusters, \textit{rbf} affinity, and \textit{$gamma$} equal to 0.01 ; clustering was done in the space of Morgan fingerprints expressed as bit-vectors. Skeletonized representation of the onium data set, i.e., Mapper graph, was constructed as follows: each disjoint subset in a level set (a cluster) was represented by a node; two nodes were linked if the corresponding clusters overlapped. Two Mapper clusters can overlap because the filter function intervals and respective level sets are allowed to overlap. 

Missing data appear in a Mapper graph as multiple connected components, loops, and branches, aka \textit{flares}. Mapper graph reveals topological features of the data set that exist at some specific scale determined by the number of the intervals covering the range of the filter function and by clustering parameters. One has to construct a family of Mapper graphs to capture evolution of topological features over multiple scales in the same exhaustive manner as PH does. As a result, Mapper is less burdensome computationally and its output is more conducive to the human analysis that associates topological features with specific data points. At the same time, Mapper offers essentially unbounded flexibility in the selection of its parameters, from selection of the filter function, to the number and overlap of filter function covering intervals, to the level set clustering parameters. The selection is frequently driven by usefulness of the produced insights rather than some numerical measure of Mapper quality. We chose parameters that resulted in a simple Mapper graph showcasing structural elements associated with lacunae in the underlying data (see \ref{sec:results}) and enabling construction of a practical example of the generative data completion.

\subsection{Scaffold-constrained generator of molecular structures.}
\label{subsec:generative}
With the advances of deep machine learning many novel approaches for molecule discovery have been proposed in recent years  \cite{elton2019deep}. While many of them show promising results, one unresolved issue is the large number of molecules they generates, of which many don't contain the critical substructures. To solve this, several recent studies propose scaffold-constrained generation of molecules where definition of the ``scaffold'' can be quite flexible \cite{li2018multi,langevin2020scaffold,C9SC04503A}. 

Our method is based on the work presented by Langevin et al. \cite{langevin2020scaffold}. The approach uses a SMILES-based Recurrent Neural Network (RNN) model \cite{rumelhart1986learning,hochreiter1997long} for molecule generation. The approach extends the standard RNN generative model using a modified sampling procedure to achieve scaffold-constrained generation. The standard RNN generative models generate SMILES sequentially, by modeling the conditional probability distribution over SMILES tokens, given input partial SMILES string. Or simply, given a set of tokens from a tokenized SMILES string, the RNN model is able to calculate the conditional probability of what will be the next token in the sequence. After the conditional probability distribution is learnt, the  sampling can be achieved by initializing the sequence with a GO SMILES token, and then sampling tokens sequentially. The approach proposed by \cite{langevin2020scaffold} extends this standard RNN model to perform scaffold-constrained sampling. The model follows a given SMILES scaffold, and only samples new tokens when when an open position of the SMILES is reached. The approach considers three different types of open positions: (i) open positions at branching points of the molecular graph; (ii) open positions in linkers (that link different cycles of the molecule); (iii) constrained choices, where the position is open but the number of possibilities is already limited within the molecule discovery project.

The conditional probability distribution was learnt from the data set of onium cations that had an artificially created lacuna (see Figure \ref{fig:mapper and generative_scaffolds}A and related description in section \ref{sec:results}). Two versions of the generator were trained \cite{olivecrona2017denovo}. In the first version, referred to as \textit{Tanimoto}, the scores of the structures generated during the training phase were based on their Tanimoto similarity to the set of query structures. In the second version, referred to as \textit{Lens}, the scores were computed as the error function between the predicted values of the anomaly scores of the generated structures, and a value of the anomaly score associated with the target level set on Mapper graph. Therefore, the \textit{Tanimoto} generator was expected to sample sulfoniums in vicinity of the scaffolds provided at the sampling stage, and the \textit{Lens} generator was expected to populate a particular level set of the Mapper graph with sulfoniums derived from the provided scaffolds. Scaffold decomposition of oniums was performed by means of \textit{RDKit} \cite{rdkit} implementation of Murco fragmentation that reduces molecular structure to the contiguous ring systems and linked rings. This approach is particularly attractive in the case of iodonium and sulfonium cations that are rich in aromatic systems. During the sampling phase, the identified reference scaffolds were augmented with placeholder symbols, \textit{($*$)} in SMILES notation, at random positions to indicate where the scaffold expansion should occur. Generator sampling was initiated with the placeholder scaffolds.

\section{Results}
\label{sec:results}
\begin{figure} 
    \centering
    \includegraphics[width=1.0\textwidth]{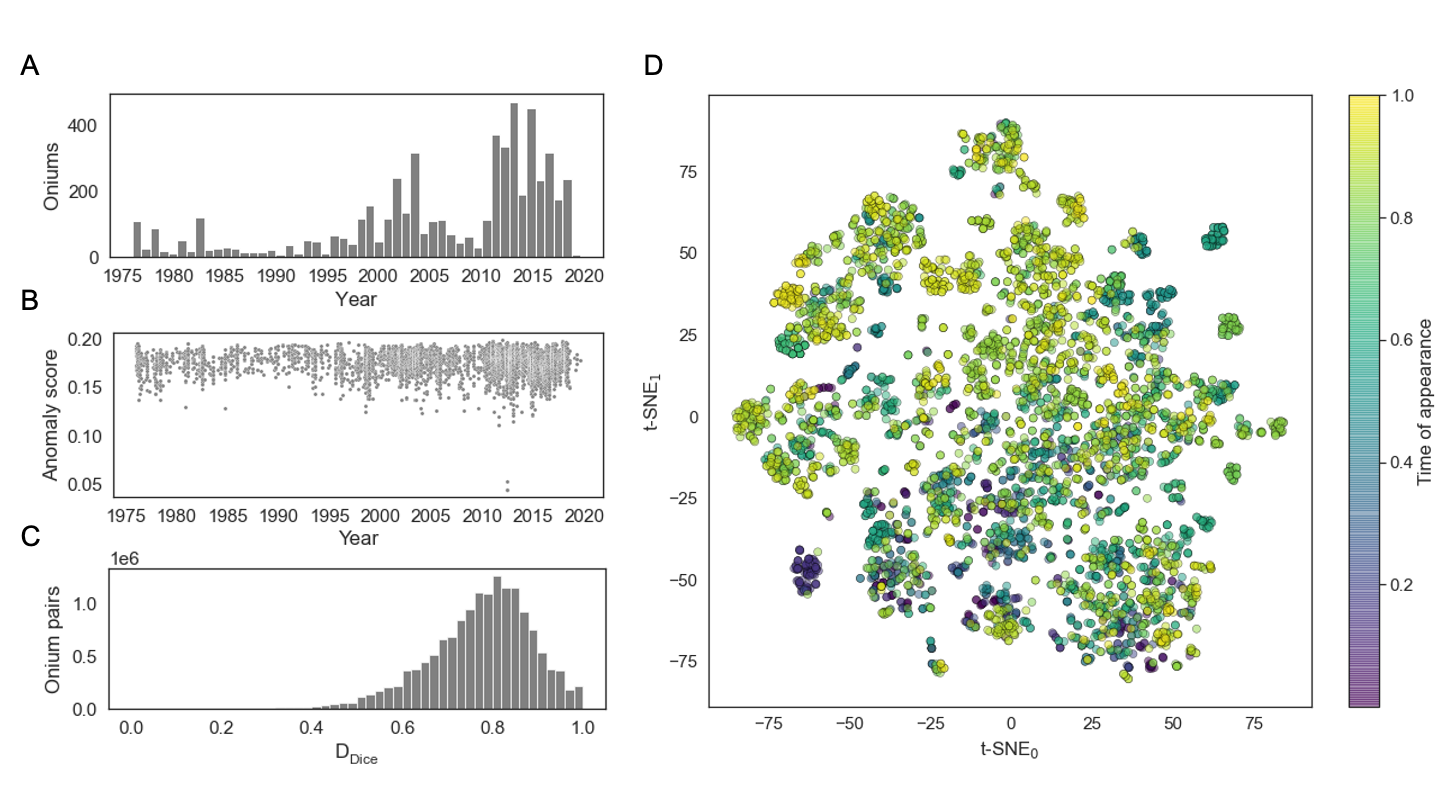}
    \caption{\label{fig:data_representation}Historical data set of onium cations extracted from USPTO patents. \textbf{Panel A}: counts of onium cations appearance in the patents. \textbf{Panel B}: anomaly score cations along the data set timeline. \textbf{Panel C}: histogram of pair-wise Dice distances ({$D_{Dice}$}) over the data set. \textbf{Panel D}: t-SNE manifold recovered from the Dice distance matrix. Color represents relative time on the data set timeline when the cations appeared in the patents.}
\end{figure}

The visualization of the constructed data set of onium cations along with analytics relevant to the historical development of the data are shown in Figure \ref{fig:data_representation}. 
Histogram in Fig. \ref{fig:data_representation}A captures historical pattern of onium cations appearance in patents between 1976 and 2019. Anomaly score of the cations in the entire data set is mapped out along the timeline of the data set accumulation in Fig. \ref{fig:data_representation}B. We notice that the cations with lower-end anomaly scores are present throughout the entire historical progression and are not associated with any specific period of time; however, there is some broadening of the range of the anomaly scores over last decade. Together with the absence of clear outliers in the data set, we interpret this as an indication that inventorship in the respective chemical domains was not driven by the discovery of outliers. 

For a cation to make its way into a patent as the intellectual-property-generating component, it has to be novel and functionally non-obvious at the time of the patent filing. This consideration renders unlikely an incremental evolution of the onium structures within the data set, aside from a) possible presence of a ``contaminating'' subset of oniums (see discussion in subsection \ref{subsec:data}), and b) the known practice in patent writing to include extensive boilerplate sections covering close analogues of the molecules that constitute the actual invention. Indeed, the mode of the histogram of pair-wise Dice distances (Figure  \ref{fig:data_representation}C) corresponds to the Dice distance 0.8, indicating evolution of the data set of patent-worthy molecules via addition of predominantly dissimilar structures. 

\begin{figure} 
    \centering
    \includegraphics[width=1.0\textwidth]{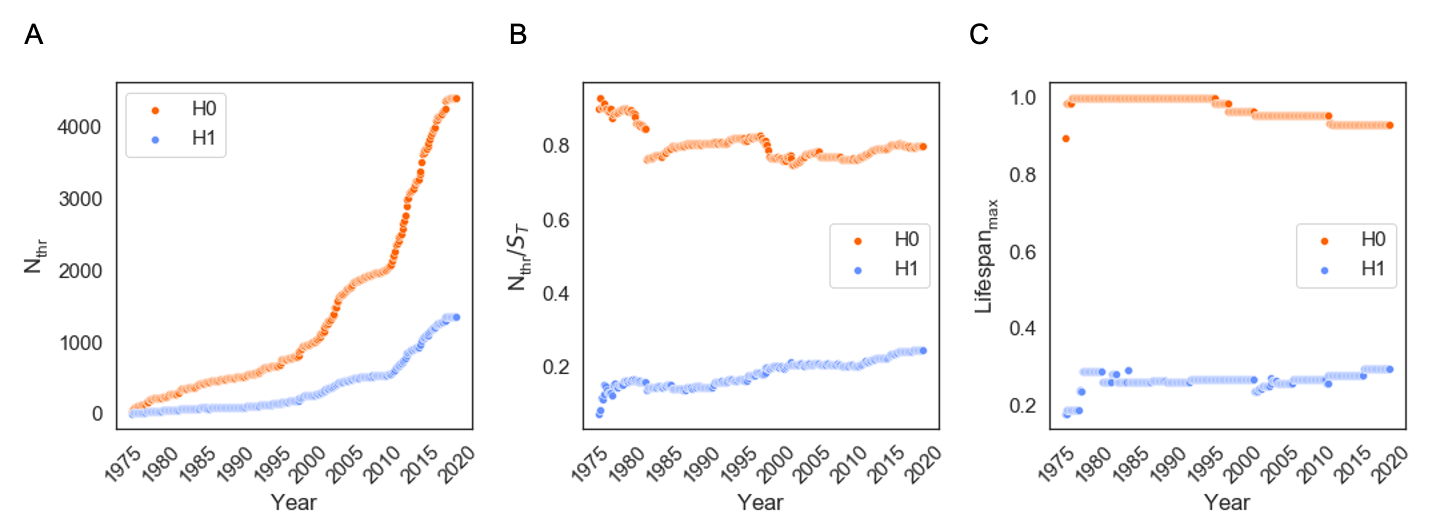}
    \caption{\label{fig:persistent_homology_stats} Summary of the PH results. \textbf{Panel A}: absolute number of topological features \textit{$N_{thr}$} with lifespan above 0.01 threshold in cumulative point clouds of oniums; homology degree 0 (\textit{H0}) corresponds to gaps and homology degree 1 (\textit{H1}) corresponds to loops. \textbf{Panel B}: number of topological features with lifespan above 0.1 threshold relative to the size \textit{$S_T$} of the cumulative point cloud at the corresponding cutoff time \textit{T}. \textbf{Panel C}: maximum lifespan of the topological features in the cumulative point clouds.}
\end{figure}

Finally, the low-dimensional t-SNE representation of the molecular point cloud indicates appreciable non-uniformity (Figure \ref{fig:data_representation}D) which would be typically elucidated via clustering analysis. At this point, we are not concerned with the presence of the clusters, but with the structure of lacunae as an indicator of non-uniformity and, ultimately, missing data. Color map in Figure \ref{fig:data_representation}D corresponds to the relative time on the timeline of the data set development when as subset of oniums appeared. We discretized the timeline of the data accumulation and constructed a time-series comprised of 319 molecular point clouds where each instance of a point cloud comprises molecules that appeared in the data set by a specific time cutoff between 1976 and 2019. PH computations were carried out for each of these cumulative point clouds. Figure \ref{fig:persistent_homology_stats}A shows that the number of topological features with degree 0 and 1 (\textit{$H_0$} and \textit{$H_1$}, respectively) is growing along the data set timeline. The plots track the features with the lifespan above 0.1 threshold, but the growth is observed for other threshold values, including no threshold imposed. Here, the scale of the features (and lifespan) is simply the Dice distance over the point cloud. The number of the detected topological features and the size of the respective cumulative point cloud show high degree of correlation, with Pearson coefficient 1.0 and p-value 0.0 for both feature degrees; their ratio shows a pattern of alternating slow growth periods with step-like declines for \textit{$H_0$} and slow growth over time for \textit{$H_1$} (Figure \ref{fig:persistent_homology_stats}B). Finally, Figure \ref{fig:persistent_homology_stats}C shows that the maximum lifespan of the topological features in cumulative point clouds remains essentially constant over time. For \textit{$H_0$} the mean of the maximum lifespan over the data set history is 0.97, and for \textit{$H_1$} the mean is 0.26. 

\begin{figure} 
    \centering
    \includegraphics[width=1.0\textwidth]{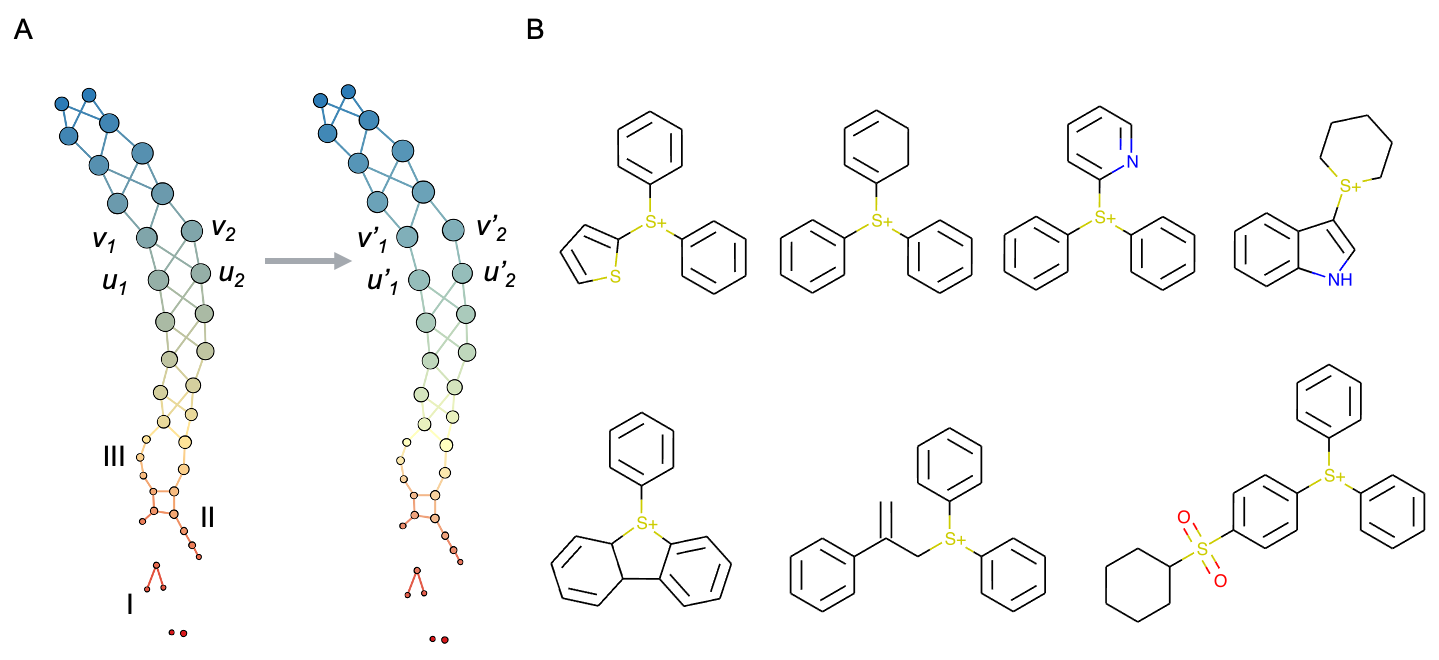}
    \caption{\label{fig:mapper and generative_scaffolds} Data-completion driven by topological features recovered using Mapper. \textbf{Panel A}: Mapper graph, a skeletonization of the molecular point cloud of oniums. Color represents levels of the Isolation Forest anomaly score of onium cations: the abnormal character is increasing from higher to lower values (from blue to red colors, cf. Fig.\ref{fig:data_representation}). The Mapper graph of the historical data is modified to create an artificial lacuna: we removed the samples at the intersections of \textit{$u_1$} with \textit{$v_2$} and \textit{$u_2$} with \textit{$v_1$}, with new sets \textit{$u'_1$}, \textit{$u'_2$}, \textit{$v'_1$}, and \textit{$v'_2$} forming a loop.  \textbf{Panel B}: reference sulfonium scaffolds associated with nodes \textit{$u'_1$}, \textit{$u'_2$}, \textit{$v'_1$}, and \textit{$v'_2$} that were used to constrain the generative procedure repairing the artificially created lacuna on the modified Mapper graph.} 
\end{figure}

PH offers summarization and featurization of the historical development of the onium data set that accounts for the missing data rather than present data. Technically, it is possible to track each individual topological feature revealed in this manner to specific cations in the molecular point cloud. The difficulty, however, is in the exhaustive nature of the PH computations. Mapper offers comparatively simpler approach to the task of finding individual topological features indicative of the lacunae in the data. The Mapper graph of the entire data set is shown in Figure \ref{fig:mapper and generative_scaffolds}A (left-hand side). Color of the nodes in Mapper graph encodes anomaly score, transition from the blue to the red colors indicates increasingly anomalous character. Mapper graph has multiple features associated with lacunae in the data at the scales that correspond to the selected parameters of the analysis. In Figure \ref{fig:mapper and generative_scaffolds}A one can see several connected components (feature I), flairs (feature II), and a loop (feature III). This graph is already suitable for data completion - of course, there is no indication what exactly is missing, which would complicate post-completion analysis. We overcome this hurdle by creating an artificial lacuna in the data set. By construction, edges on Mapper graph correspond to the intersections of the clusters of molecules that are represented by nodes. We picked four nodes \textit{$u_1, u_2, v_1$} and \textit{$v_2$} on the original Mapper graph (Figure \ref{fig:mapper and generative_scaffolds}A, left-hand side), retrieved corresponding sets of oniums and computed their intersections, and removed 486 oniums corresponding to the edge (\textit{$u_1, v_2$}) and 359 oniums corresponding to the edge (\textit{$u_2, v_1$}). The updated Mapper graph with an artificial lacuna is shown on the right-hand side of Figure \ref{fig:mapper and generative_scaffolds}A. The data completion task, therefore, is to generate new oniums that create edges (\textit{$u'_1, v'_2$}) and (\textit{$u'_2, v'_1$}).

Generative data completion proceeded as described in subsection \ref{subsec:generative}. The \textit{Tanimoto} generator was trained with the query structures  comprising 30 sulfonium-containing scaffolds simultaneously present among withheld cations and among remaining cations associated with nodes \textit{$u'_1, u'_2, v'_1$} and \textit{$v'_2$}, but nowhere else in the data set. The \textit{Lens} generator was trained with the target anomaly score equal to the average of the highest anomaly score among the cations associated with nodes \textit{$u_1$} and \textit{$u_2$} (0.1758) and the lowest anomaly score among the cations associated with nodes \textit{$v_1$} and \textit{$v_2$} (0.1707). Referring to the procedure of Mapper graph construction (subsection \ref{subsec:tda}), these are the bounds of the overlap region between two intervals of the filter function that contain clusters represented by nodes \textit{$u'_1, u'_2, v'_1$} and \textit{$v'_2$}; these nodes can only have links if there are molecules ``populating'' this overlap region. Reference sulfonium scaffolds shown in Figure \ref{fig:mapper and generative_scaffolds}B were used to construct 65 placeholder scaffolds that initiated sampling from the trained generative models.

\begin{figure} 
    \centering
    \includegraphics[width=1.0\textwidth]{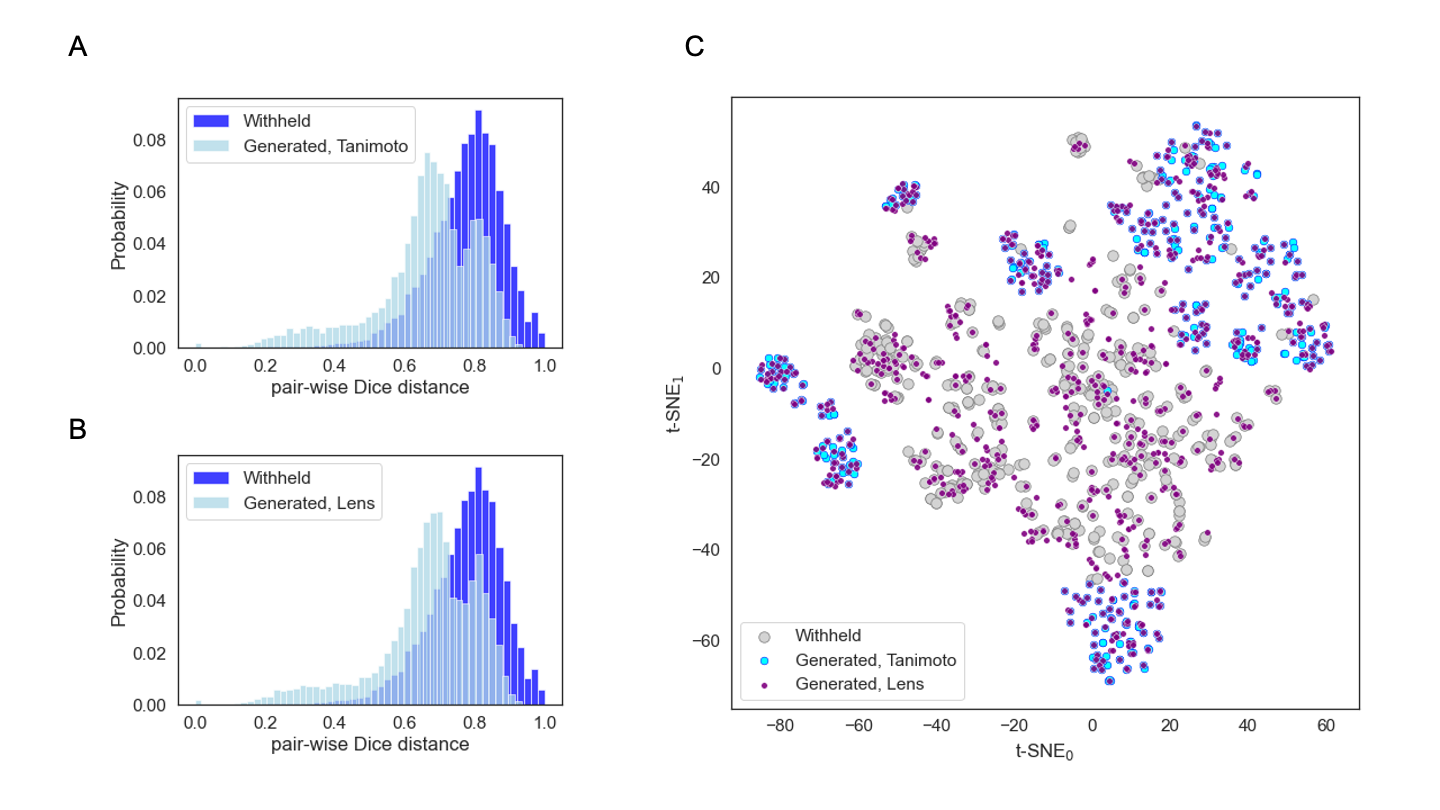}
\caption{\label{fig:completion_comparison}Comparison of the samples produced by generative procedure aiming to complete artificially created lacunae in the data set with the withheld sample. \textbf{Panel A}: Probabilities of Dice distances within the withheld sample and within the sample generated by a RL agent trained with \textit{Tanimoto} scoring function. \textbf{Panel B}: same, RL agent trained with \textit{Lens} scoring function. \textbf{Panel C}: t-SNE embedding of the withheld sample, and both generated samples after filtering by anomaly score.}
\end{figure}

We harvested approx. 3.3K oniums from the \textit{Tanimoto} generator and approx. 3.6K oniums from the \textit{Lens} generator. The output of both generative models was filtered to select sub-samples with anomaly scores in the range (0.1707,0.1758) - we discussed the relevance of this filter function interval earlier. There were 406 and 434 cations in the filtered samples derived from \textit{Tanimoto} and \textit{Lens} generators, respectively.  The probabilities of pair-wise Dice distances in the filtered samples are shown in Figures \ref{fig:completion_comparison}A and B along with the probability of pair-wise Dice distances in the withheld set of onium cations. The latter shows a single mode close to 0.8, similarly to the distribution over the entire data set in Figure \ref{fig:data_representation}C. In contrast, both generated samples show bi-modal distributions with more prominent features close to 0.7 and weaker features close to 0.8. Low-dimensional manifold embedding of all three sets of cations (Figure \ref{fig:completion_comparison}C) helps rationalize this result. Both generated samples exhibit clustering patterns attributable to the bias of the generative model towards populating vicinity of the reference scaffolds. Unlike the \textit{Tanimoto} sample, the \textit{Lens} sample includes cations located between such clusters. Withheld data and \textit{Tanimoto} sample appear to occupy complementary regions with respect to each other; \textit{Lens} sample covers the union of those regions. Of course, these are qualitative assessments based on a low-dimensional representation of the data that can introduce appreciable artifacts. 

Finally, we evaluated if the generated sulfonium cations completed the artificially created lacuna in the data set. In addition to the filtered samples we created their downsampled versions where we removed cations with large number of neighbors within 0.0 to 0.6 range of Dice distances. This downsampling strategy was intended to bring the shape of the distance distribution in the generated data closer to the original data. The original data set minus withheld cations was updated with the samples of the generated cations and the updated Mapper graph was constructed. Table \ref{tab:datacompletions} summarizes results of the generative completion on the Mapper graph. The implemented generative workflow successfully repaired the artificially created lacuna in the onium data set by reconstructing the removed links. Addition of the downsampled cations sourced from either version of the generator completed both artificially removed links, \textit{$u'_1, v'_2$}) and (\textit{$u'_2, v'_1$}) (see Figure \ref{fig:mapper and generative_scaffolds}A, right-hand side). Unexpectedly, addition of the cations that were not downsampled completed only one link, \textit{$u'_1, v'_2$}) for both sample sources. This behavior was attributed to the peculiarity of Mapper graph construction: addition of new molecules to the intersection of the relevant level sets according to the value of the filter function, i.e., anomaly score, guarantees that \textit{some} links will be created; however, assignment of the new molecules to the clusters within the level sets depends on topology of the set of new molecules and specifics of the clustering algorithm. Without downsampling, presence of close neighbors pushed clustering algorithm to assign new molecules to two clusters instead of four, creating only one link. In addition, spectral clustering does not ``learn'' a model that can be instantiated and re-used on the new data to force their assignment to the pre-existing clusters.  

\begin{table}
 \caption{Results of the generative completion on Mapper graph} \label{tab:datacompletions}
  \centering
  \begin{threeparttable}
  \begin{tabular}{llll}
    \toprule

    Generator     & Restored links/New cations  \tnote{A}      & Restored links/New cations \tnote{B} & Restored links/New cations \tnote{C} \\
    \midrule
    Tanimoto & (\textit{$u'_1, v'_2$})/406  & (\textit{$u'_1, v'_2$}), (\textit{$u'_2, v'_1$})/375 & (\textit{$u'_1, v'_2$}), (\textit{$u'_2, v'_1$})/238  \\
    Lens     & (\textit{$u'_1, v'_2$})/434  & (\textit{$u'_1, v'_2$}), (\textit{$u'_2, v'_1$})/392 & (\textit{$u'_1, v'_2$}), (\textit{$u'_2, v'_1$})/254   \\

    \bottomrule
  \end{tabular}
  \begin{tablenotes}
  \item[A] Selection by anomaly score.
  \item[B] Same as A, removed cations with more than 300 nearest neighbors.
  \item[C] Same as A, removed cations with more than 200 nearest neighbors.
  \end{tablenotes}
  \end{threeparttable}
 \end{table}

\section{Discussion}
\label{sec:discussion}

In this contribution, we used two forms of TDA, PH and Mapper, to analyze a historical data set of patentable onium cations and a) express development of the data set in terms of lacunae in the data, b) demonstrate that lacunae can be identified in a manner conducive to hypothesis generation, and c) drive generative modeling to produce new onium cations that complete a specific lacuna. PH calculations over the timeline of the data set development demonstrated abundance of topological features associated with gaps and loops in the data. Although the full characterization of the historical progression of the data set in terms of 3- or higher-dimensional voids turned out to be computationally challenging, the partial analysis indicated presence of these features (results not included). Our results show that the number of gaps and loops strongly correlates with the data set size (Figures \ref{fig:persistent_homology_stats}A and B) and behaves like an extensive property of the data set. The tentative interpretation of this finding is that historical development of the pertinent technological fields was not driven by elimination of lacunae. This might be a consequence of the requirement for non-obvious/novel character of patentable materials and/or lack of any particular effective ``strategy'' of expansion over the set of potential materials, at least as far as elimination of the empty spaces in the data is concerned.

Analysis of the maximum lifetimes of the topological features (Figure (\ref{fig:persistent_homology_stats}C)) did not reveal a pattern of historical completion of lacunae, either. There is a possibility, however unlikely, that elimination of prominent lacunae coincided with appearance of new once. We notice a slight difference in historical trends for the features of degree 0 and 1 in Figure \ref{fig:persistent_homology_stats}C, where the former shows slight decline and the latter shows slight growth. It remains to be seen if the revealed historical patterns of PH results are specific to the intellectual property generation, especially in the view of predominantly dissimilar nature of the cations populating the data set (Figure \ref{fig:data_representation}C), which is counter to the expectation of an incremental historical evolution of the data.

It is clear that our data set is rich in lacunae. The question remains, is the content of these lacunae useful? In order to start exploring the answer, we need to bridge TDA capabilities with generative modeling. Here, Mapper offers a complementary capability to the exhaustive enumeration of topological features in PH computations. Mapper graph, a skeletonized representation of the high-dimensional molecular point cloud (Figure \ref{fig:mapper and generative_scaffolds}A), clearly conveys the presence of individual topological features and enables direct access to the original data associated with these features. These are not mere conveniences; the level of human ownership, stakeholding, and decision-making in materials discovery creates a demand for the workflows that facilitate capture and injection of expert input into the hypothesis generation \cite{EITL:2020}. The operational decisions that we made initiating the generative phase included selection of the reference scaffolds (Figure \ref{fig:mapper and generative_scaffolds}) and transformation of the reference scaffolds into placeholder scaffolds (subsection \ref{subsec:generative}). These decisions are necessarily human-driven, they are likely to follow expert-specific logic and lead to varying results.

There is a multitude of generative modeling algorithms that can be plugged in the presented approach. There are a lot of model-specific factors at play. The algorithm that we adopted is reinforcement learning for the sequence generation. It turned out that the model had difficulties with correct enumeration of multiple cycles frequently present in cationic oniums; large fraction of such instances was corrected via postprocessing of the generated strings. We chose to train the prior using our data instead of taking advantage of the prior pretrained on the content of a chemical database. This decision helped to avoid expanding sulfonium scaffolds with fragments of drug-like compounds, but it noticeably decreased the number of unique molecules harvested at the sampling phase. The sampled sulfoniums had a broad distribution of the anomaly scores, even with the \textit{Lens} generator trained to hit a narrow interval of values - as a result, the number of new cations added to the data set was almost an order of magnitude lower than the number of sampled cations. 

It appears that the choice of the scoring function during the generator training phase effectively controlled the balance between exploration and exploitation during the sampling phase (Figure \ref{fig:completion_comparison}). \textit{Lens} generator produced more appealing results in terms of the internal structure of the generated sample. Of course, both generators demonstrated ability to fill in the artificially created lacuna and complete links on Mapper graph. A systematic investigation and design of the scoring function that offer higher level of control over the topology of the generated samples would be the most consequential extension of the presented work. 

Mapping out patterns of development of the historical data sets in materials domain is important for several reasons. Materials discovery is an issue of finding optimal discovery strategies as much as it is an issue of materials modeling, physical or data-centric. Historical data tell the story of the successes, they need to tell the story of the efficiency, realistic ratios between efforts and returns, and missed opportunities. Explicit characterization of lacunae in the historical data offers a route to such insights. In a more practical, data-centric context of the discovery, it points us to the deficiencies of the data sets that have a real chance of misguiding efforts informed by statistical models. In material science, the cost of experimentation is high and enrichment of the experiments with meaningful candidates is a challenge that has yet to be overcome. A specialized form of this challenge relates directly to the expansion of the generative modeling to the materials discovery domain. There is no doubt that it \textit{can} be a useful approach; working out the means of robust control over \textit{naive} generative models and taming their creativity will help to cover the gap between the computational and experimental sides of the materials discovery \cite{AMDARC:2019}.

\bibliographystyle{unsrt}  
\bibliography{references}  

\end{document}